# The Evolution of Sex through the Baldwin Effect

Larry Bull

Department of Computer Science & Creative Technologies

University of the West of England

Bristol BS16 1QY, U.K.

+44 (0)117 3283161

Larry.Bull@uwe.ac.uk

## Abstract

This paper suggests that the fundamental haploid-diploid cycle of eukaryotic sex exploits a rudimentary form of the Baldwin effect. With this explanation for the basic cycle, the other associated phenomena can be explained as evolution tuning the amount and frequency of learning experienced by an organism. Using the well-known NK model of fitness landscapes it is shown that varying landscape ruggedness varies the benefit of the haploid-diploid cycle, whether based upon endomitosis or syngamy. The utility of pre-meiotic doubling and recombination during the cycle are also shown to vary with landscape ruggedness. This view is suggested as underpinning, rather than contradicting, many existing explanations for sex.

1. Introduction

Whilst a number of explanations for various aspects of the evolution and maintenance of eukaryotic sex have been presented, none gives a unifying view of the wide variations in the process seen in nature. Sex is here defined as successive rounds of syngamy and meiosis in a haploid-diploid lifecycle. This paper suggests that the emergence of a haploid-diploid cycle enabled the exploitation of a rudimentary form of the Baldwin effect [1][21][25] and that this provides an underpinning explanation for all the observed forms of sex. The Baldwin effect is here defined as the existence of phenotypic plasticity that enables an organism to exhibit a significantly different (better) fitness than its genome directly represents. Over time, as evolution is guided towards such regions under selection, higher fitness alleles/genomes which rely less upon the phenotypic plasticity can be discovered and become assimilated into the population (see [31] for a recent overview). Alongside neural processing, the Baldwin effect has been connected to other aspects of organisms, such as the immune system [14].

Hinton and Nowlan [15] were the first to investigate the Baldwin effect, showing that enabling genetically specified neural networks to alter inter-neuron connections randomly during their lifetime meant the evolutionary system was able to find an isolated optimum, something the system without learning struggled to achieve. That is, the ability to learn "smoothed" the fitness landscape into a unimodal hill/peak. They also found that over time more and more correct connections became genetically specified and hence less and less random learning was necessary; the evolutionary process was guided toward the optimum by the learning process. Belew (eg, [2]) added the Baldwin effect via backpropogation to his work on the evolution of neural networks for various classes of problem, finding that the search process was greatly improved. Stork and Keesing [29] then showed how both the frequency with which learning was applied and the number of connection weight adjustment iterations used on each learning cycle impacted upon the benefit gained. Their finding was generalized in [6] where it was shown how *the most beneficial frequency and amount of learning varies with the ruggedness of the underlying fitness landscape*. This paper shows how various aspects of sex can be seen as mechanisms through which to alter either the frequency and/or amount of learning in a haploid-diploid lifecycle to match the underlying ruggedness of the organism's fitness landscape.

As discussed in [24, p150] the first step in the evolution of eukaryotic sex was the emergence of a haploid-diploid cycle, probably via endomitosis, before simple syngamy. Cleveland [10] was first to suggest that organisms may become diploid by a variation in mitosis to maintain the genome copy, ie, endomitosis. Syngamy, the fusion of two independent genomes, probably emerged thereafter. The subsequent emergence of isogamy, ie, mating types, is not considered in this paper. Under both scenarios, a previously haploid cell became diploid. A number of explanations have been presented for why a diploid, or increasing ploidy in general, is beneficial, typically based around the potential for "hiding" mutations within extra copies of the genome (eg, see [26] for an overview). A change in ploidy can potentially alter gene expression, and hence the phenotype, even if no mutations occur between the lower and higher ploidy states - through epigenetic mechanisms, through rates of changes in gene product concentrations, no or partial or co-dominance, etc. (eg, see [9]). In all cases, whether the diploid is formed via endomitosis or syngamy, *the fitness of the cell/organism is a combination of the fitness contributions of the composite haploid genomes*. If the cell subsequently remains diploid and reproduces asexually, there is no scope for a rudimentary Baldwin effect. However, if there is a reversion to haploid cells under meiosis, there is potential for a mismatch between the utility of the haploids compared to that of the polyploid; individual haploids do not contain all of the genetic material over which selection operated. That is, the effects of genome combination can be seen as a simple form of phenotypic plasticity for the individual haploid genomes before they revert to a solitary state and hence the Baldwin effect may occur.

This paper begins by revisiting the results presented in [6] using the NK model [17], before extending the model to consider the evolution of various aspects of eukaryotic sex in a single celled organism in light of its findings.

## 2. The NK Model

Kauffman and Levin [17] introduced the NK model to allow the systematic study of various aspects of fitness landscapes (see [16] for an overview). In the standard model, the features of the fitness landscapes are specified by two parameters: $N$, the length of the genome; and $K$, the number of genes that has an effect on the fitness contribution of each (binary) gene. Thus increasing $K$ with respect to $N$ increases the epistatic linkage, increasing the ruggedness/complexity of the fitness landscape. Kauffman [16] shows that the increase in epistasis increases the number of optima, increases the steepness of their sides, and decreases their correlation.

The model assumes all intragenome interactions are so complex that it is only appropriate to assign random values to their effects on fitness. Therefore for each of the possible $K$ interactions a table of $2^{(K+1)}$ fitnesses is created for each gene with all entries in the range 0.0 to 1.0, such that there is one fitness for each combination of traits (Figure 1). The fitness contribution of each gene is found from its table. These fitnesses are then summed and normalized by $N$ to give the selective fitness of the total genome.

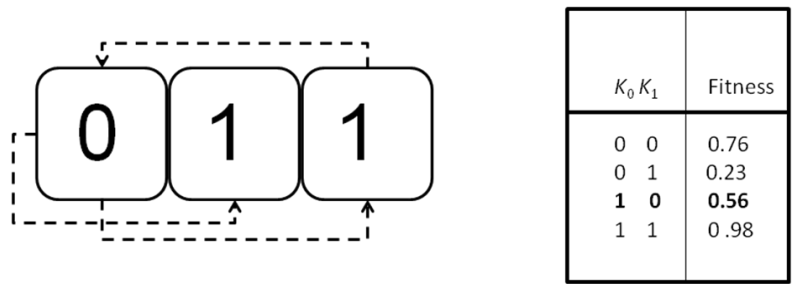

Figure 1. An example NK model ($N=3$, $K=1$) showing how the fitness contribution of each gene depends on $K$ random genes (left). Therefore there are $2^{(K+1)}$ possible allele combinations per gene, each of which is assigned a random fitness. Each gene of the genome has such a table created for it (right, centre gene shown). Total fitness is the normalized sum of these values.

Kauffman [16] used a mutation-based hill-climbing algorithm, where the single point in the fitness space is said to represent a converged species, to examine the properties and evolutionary dynamics of the NK model. That is, the population is of size one and a species evolves by making a random change to one randomly chosen gene per generation. The "population" is said to move to the genetic configuration of the mutated individual if its fitness is greater than the fitness of the current individual; the rate of supply of mutants is seen as slow compared to the actions of selection. Following [6], a very simple (random) learning process to enable phenotypic plasticity can be added to evolution by allowing a new individual to make a further $L$ (unique) mutations after the first. If the averaged fitness of this "learned" configuration and that of the first mutant is greater than that of the original, the species is said to move to the *first* mutant configuration but assigned the *averaged fitness* of the two configurations. All results reported in this paper are the average of 10 runs (random start points) on each of 10 NK functions, that is 100 runs, for 50,000 generations. Here $0 \leq K \leq 15$ and $0 < L \leq 7$, for $N=20$ and $N=100$.

## 3. The Baldwin Effect in the NK Model

Figure 2 shows the performance of the Baldwin effect across a wide range of $K$ and $L$ combinations for $N=20$. For $K=0$, the unimodal case, learning shows no benefit for evolution (T-test, $p\geq0.05$, $0<L<7$) and is disruptive when applied at higher levels (T-test, $p<0.05$, $L=7$). As $K$ increases, ie, as landscape ruggedness increases, learning becomes beneficial across a wider range of $L$. When $0<K<6$, learning is either beneficial (T-test, $p<0.05$, $0<L<7$), or has no effect (T-test, $p\geq0.05$, $L=7$). Learning is always beneficial over the ranges used when $K\geq6$ (T-test, $p<0.05$). The smallest amount of learning $L=1$ is as beneficial as any other until $K>6$, when the higher levels are most beneficial (T-test, $p<0.05$, $L\geq5$). The same results are typically seen for $N=100$ (not shown), although the higher amounts of learning are not found beneficial for higher $K$. These findings support those reported in [6]: the most beneficial amount of learning varies with $K$.

As noted above, as well as the most useful amount of such random learning varying with the ruggedness of the fitness landscape, ie, $L$, it was also shown how the frequency at which the simple learning process is applied to an individual can alter the conditions under which the Baldwin effect is beneficial [6].

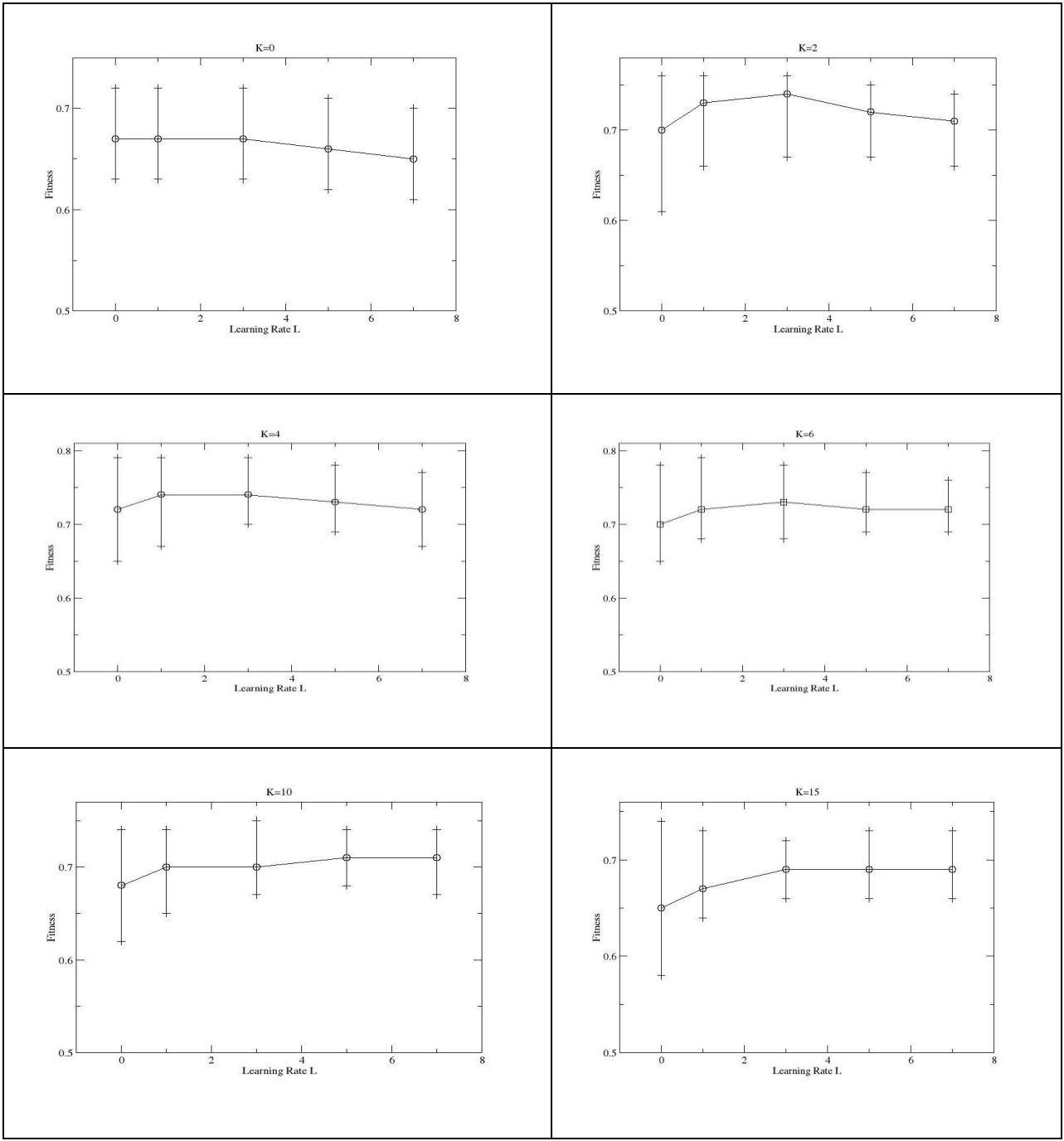

Figure 2. Performance of the Baldwin effect, after 50,000 generations, for varying amounts of learning ($L$), on landscapes of varying ruggedness ($K$) with $N$=20. Error bars show min and max values in all graphs.

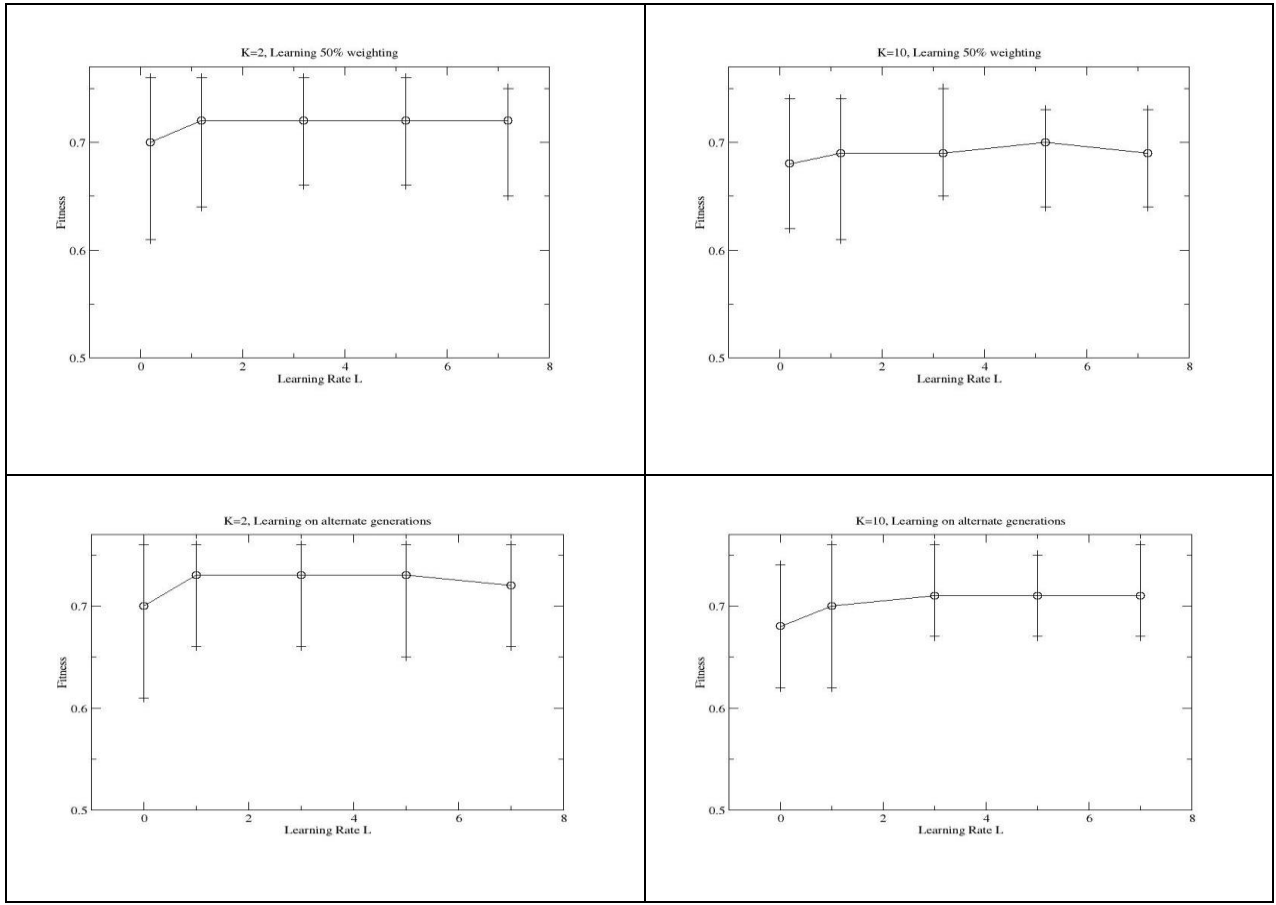

Figure 3: Performance of the Baldwin effect, after 50,000 generations, where learning occurs at different frequencies ($N=20$). The top row shows examples of learning occurring for half of the lifetime and the bottom row shows learning occurring every other generation.

Figure 3 shows examples from varying the frequency of learning both within and across lifecycles. In the previous results, the fitness of a genome was calculated as the average of its purely genetic configuration and that of the learned configuration. Thus learning can be seen to have occurred throughout the lifecycle. This can be varied such that the fitness of the learned configuration is weighted less equally to the genetic configuration: less learning. Examples of the case of learning being weighted at 50% are shown in the top row of Figure 3. Results show learning is now beneficial for all $L$ for $K=2$, with no significant change in behaviour to Figure 2 for $2<K\leq6$, whilst learning is no longer beneficial for $K>6$ (T-test, $p\geq0.05$). Figure 3 also shows examples from only allowing the original whole lifecycle learning to occur on every other generation. The results are the same as for the half lifecycle case, except there is no drop in benefit for $K>6$.

## 4. Evolution of the Haploid-Diploid Cycle: the Baldwin Effect

Whether the haploid-diploid cycle emerged via endomitosis or via a simple form of syngamy is not crucial to the basic hypothesis presented here. As noted above, explanations primarily based around mutation hiding have been given as to why a diploid state is beneficial to a haploid state. Similarly, there are explanations for the emergence of the alternation between the two states, typically based upon its being driven by changes in the environment (after [22]). If, as suggested here, the diploid state should be seen as the "learning" part of the lifecycle due to genome interactions, the results above anticipate the wide range of different haploid-diploid frequencies seen in nature. For example, most mammals have a primarily diploid lifecycle, many land plants exploit a (significant) haploid seed phase, etc. That is, as $K$ and $L$ vary, the optimal frequency with which learning occurs varies. Following [24, p150], endomitosis is assumed to have occurred first in this paper.

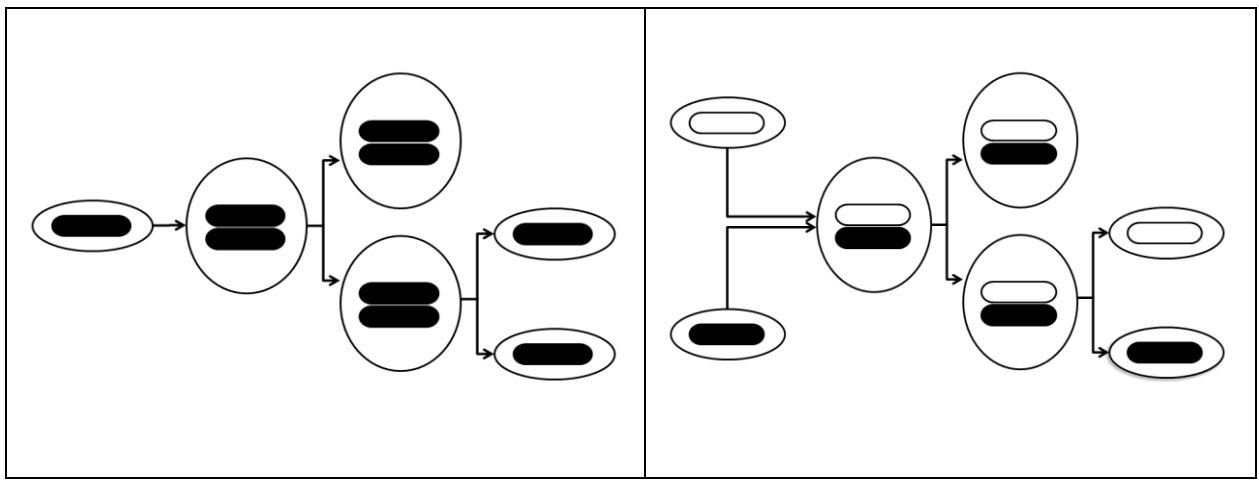

Figure 4: The endomitosis (left) and syngamy (right) processes explored here (after [Maynard Smith & Szathmary, 1995]).

The model described in section 2 is altered such that once the haploid genetic mutant is created, a copy is made and another gene chosen for further mutation (Figure 4). Both genomes are evaluated and, for simplicity, the fitness of the diploid is assigned as their average. If the diploid is fitter than the diploid representing the current population, the species is said to move to the new configuration. Again for simplicity, selection picks one of the two genomes of the diploid at random. For higher levels of ploidy (four and eight) explored, the copy and mutation process is repeated equally for each new genome to the required level. That is, the rounds of endomitosis can be seen as rounds of learning by the cell/organism.

Figure 5 shows examples of how a haploid-diploid cycle via endomitosis is beneficial over a purely haploid (non-learning) cycle for all $K>0$ (T-test, $p<0.05$). It can also be seen that a further round of endomitosis to a tetraploid state before meiosis provides no benefit over diploidy for any $K$ (T-test, $p\geq0.05$), except when $K=4$ (T-test, $p<0.05$), with another round to octaploidy providing no benefit for low $K$ and becoming detrimental for $K\geq6$ (T-test, $p\geq0.05$). The same behaviour was found for $N=100$ (not shown).

The case for a haploid-diploid being beneficial was predicted above since the endomitotically produced genome is the same as the $L=1$ case which was found to be beneficial for all $K>0$. That the tetraploidy case is beneficial over the haploid for higher $K$ is also anticipated by the previous results. However, whilst tetraploidy and octaploidy may be seen as the $L=3$ and $L=7$ cases respectively, they are subtly different. In the basic model, all $L$ random learning changes are made in *one* copy of the genome. In the polyploidy cases, each further random learning change is made in genomes copied from genomes which have already had changes made. Hence increasing ploidy both increases the distance learning can sample from the original evolution produced genome point in the fitness landscape *and* the number of learning samples. For example, two genomes have $L=1$ and one $L=2$ in the tetraploid case. Thus increasing the number of samples also appears disruptive, even for lower range $L$.

Figure 6 shows the comparative performance of the haploid-diploid cycle under endomitosis from Figure 5 to that of the equivalent simple syngamy case. In the latter, the new diploid is created either by copying and mutating one gene in each of the species' two genomes, or by copying either genome twice and then mutating each once (Figure 4). Both genomes are initialized as the same.

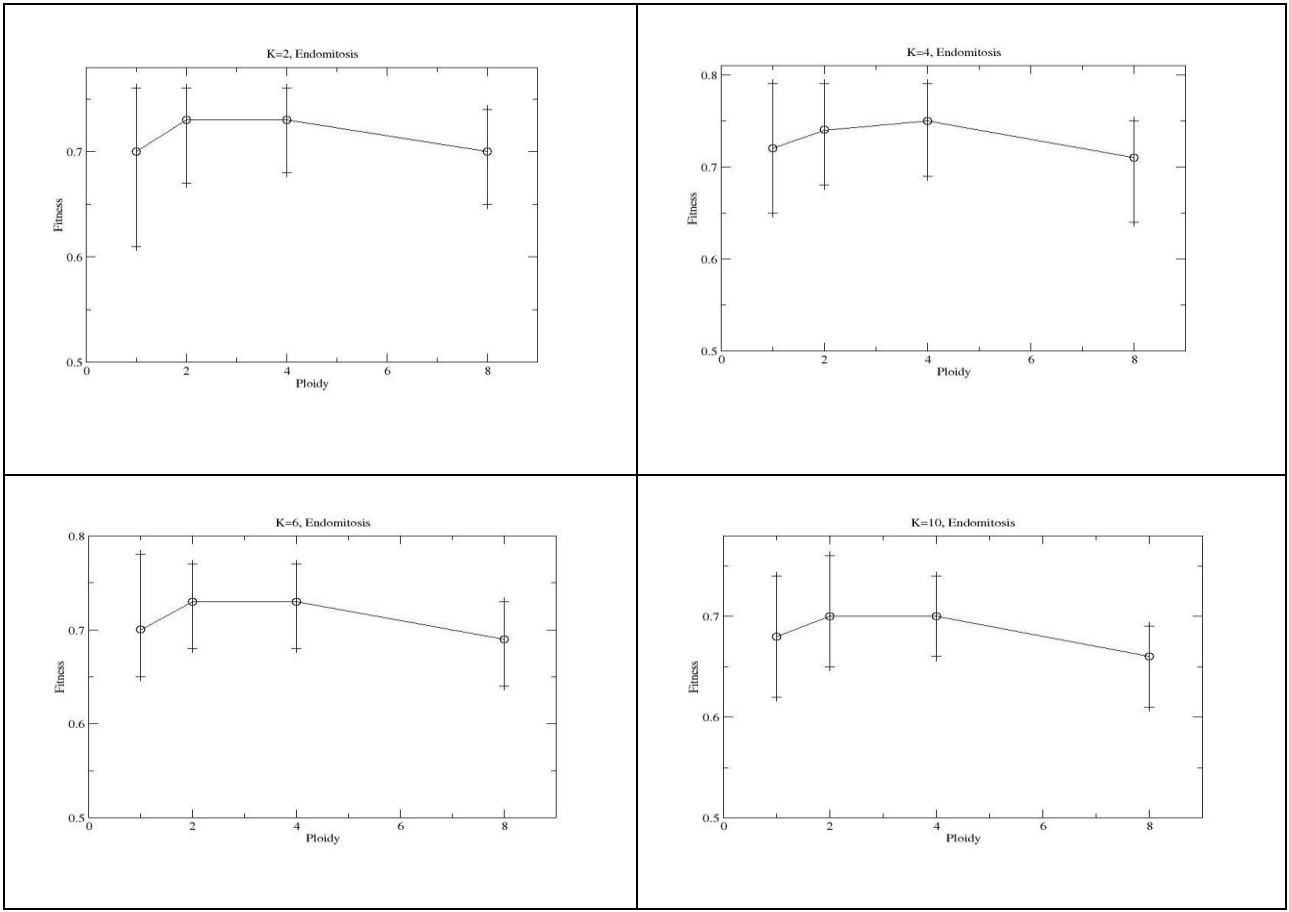

Figure 5. Performance of the Baldwin effect under endomitosis, after 50,000 generations, for varying amounts of ploidy/learning, on landscapes of varying ruggedness (*K*) with *N*=20.

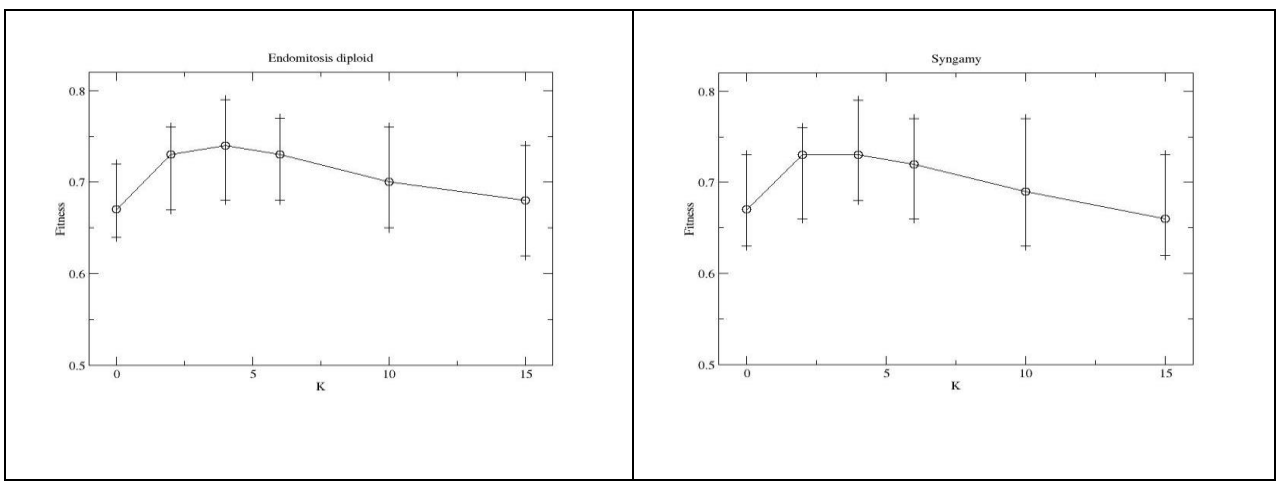

Figure 6. Comparing the performance of the Baldwin effect under endomitosis and syngamy, after 50,000 generations, on landscapes of varying ruggedness (*K*) with *N*=20.

Figure 6 shows how there is no difference between either mechanism to provide the diploid stage for $K<6$, whereafter endomitosis proves more beneficial (T-test, $p<0.05$). The reason for this difference is again due to the difference in the amount of learning occurring per cycle; the results in Figure 2 indicate a general benefit from an increased amount of learning with increasing fitness landscape ruggedness. In the endomitosis case, the learning change is added onto the genetic mutation of the first offspring genome in the second offspring genome. In the syngamy case, both genomes undergo the first genetic mutation change only. When the same genome is chosen twice to form the diploid, the syngamy case's sampling distance in the fitness landscape from the evolutionary origin is reduced in comparison to the equivalent endomitosis case (by one mutant step). When the two genomes are different in the syngamy case, this is not necessarily true, depending upon the degree of genetic diversity between the two original haploid genomes. As above, when $N=100$, the extra learning - of endomitosis – provides no extra benefit and both perform equally well for all $K$ (not shown). Comparison with an equivalent asexual diploid finds both endomitosis and syngamy more beneficial for all $K>2$ (T-test, $p<0.05$, not shown). This is also true even if all three possible diploid combinations from the two haploid genomes are evaluated per generation (T-test, $p<0.05$, not shown). As noted above, no Baldwin effect can occur.

The type of Baldwin effect working here can be seen to alter the general characteristics of the evolutionary process. In the traditional haploid view of evolution, variation operators such as mutation copy errors, gene transfers, etc., generate a new genome at a *point* in the fitness landscape for evaluation. Under the haploid-diploid cycle, the variation operators create the bounds for sampling a *region* within the haploid fitness landscape by specifying two end points, ie, each haploid genome to be partnered in the diploid. The actual position of the fitness point for the (diploid) phenotype taken from within that region then depends upon the percentage of the lifecycle the diploid state occupies - the larger, the closer to the midpoint (with all other things being equal) in the haploid landscape. Significantly, evolution assigns a single fitness value to the region of the fitness landscape the two haploid genomes delineate – evolution can be seen to be *generalizing* over the space.

As noted above, whilst the effects of the interactions between the two haploid genomes can be expected to be non-linear in nature, in the model a simpler relationship is assumed and an average fitness assigned. Figure 7 shows a very simple example of the contrast between the standard haploid genome landscape evolution view and how the haploid-diploid cycle alters the landscape with fitness contribution averaging. With the Baldwin effect the apparent fitness of the valley is potentially increased, increasing the likelihood selection will maintain such genomes within the population, thereby increasing the likelihood of the valley being crossed to the optimum. This explains the increased benefit of the haploid-diploid cycle seen above as landscape ruggedness is increased. It can also be noted that the shape of the fitness landscape *varies* based upon the haploid genomes which exist within a given population at any time and how they are paired. This is also significant since, as has been pointed out for coevolutionary fitness landscapes [7], such movement potentially enables the *temporary* creation of neutral paths, where the benefits of (static) landscape neutrality are well-established [18].

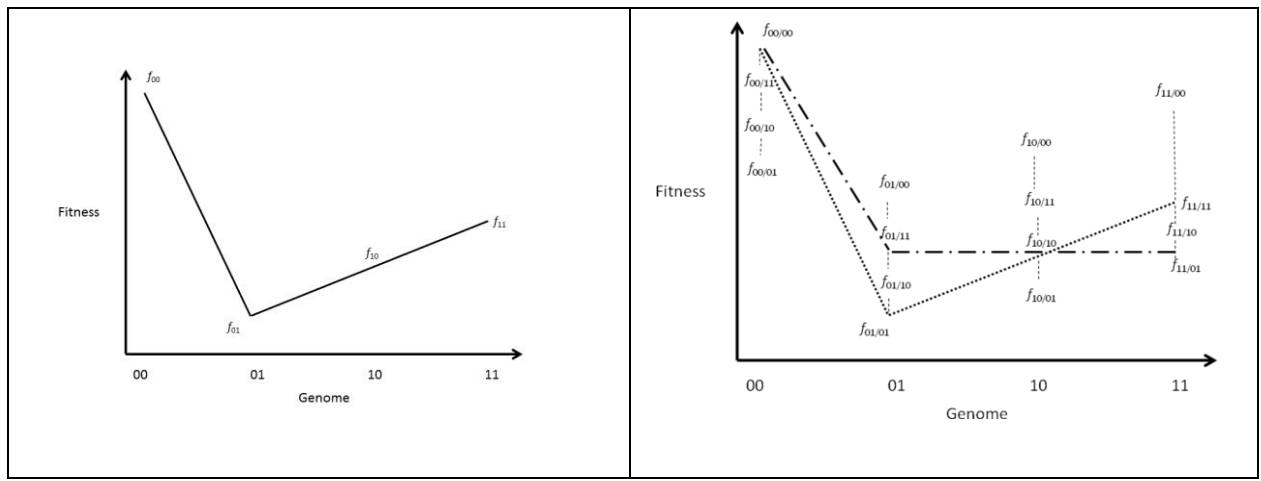

Figure 7. Comparing the fitnesses of haploid genomes under the haploid-diploid case (right) to the traditional haploid case (left), where the interaction between the two genomes in the diploid is a simple average. The broken dashed line shows the example landscape experienced by the diploids when the haploid '11' and '01' genomes always exist together in the population and the others are homozygotes. Note the loss of difference in fitness between the non-optimal genomes compared the traditional case on the left – the probability of reaching the optimum is increased.

5. Two-step Meiosis and Recombination: Altering the Amount of Learning

The few explanations as to why a form of meiosis exists which includes a genome doubling stage – the diploid temporarily becomes a tetraploid – range from DNA repair (eg, [3]) to the suppression of potentially selfish/damaging alleles (after [12]). Explanations for the recombination stage vary from the removal of deleterious mutations (eg, [19] to avoiding parasites (after [13]) (see [4] for an overview). With the Baldwin effect view proposed here, such sexual reproduction can be seen as a mechanism through which to vary the amount of learning a cell/organism can exploit during the diploid phase. The role of recombination becomes clear under the Baldwin effect view: *recombination moves the current end points in the underlying haploid fitness space which define the generalization either closer together or further apart*. That is, recombination adjusts the size of an area assigned a single fitness value, potentially enabling higher fitness regions to be more accurately identified over time. Moreover, recombination can also be seen to facilitate genetic assimilation within the simple form of the Baldwin effect. That is, the pairing of haploid genomes is seen as a "learning" step with the fitness of a given haploid affected by the allele values of its partner. If the pairing is beneficial and the diploid cell/organism is chosen under selection to reproduce, the recombination process brings an assortment of those partnered genes together into new haploid genomes. In this way the fitter alleles from the pair of partnered haploids may come to exist within individual haploids more quickly than under mutation alone.

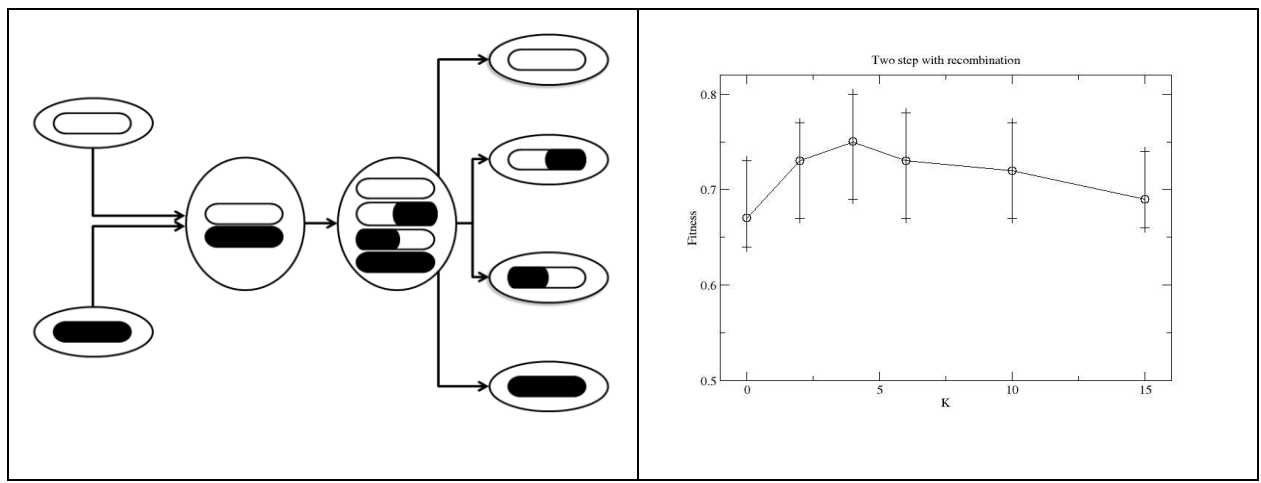

Figure 8: Two-step meiosis with recombination process (left) and its performance, after 50,000 generations, on landscapes of varying ruggedness ($K$) with $N=20$ (right).

The previous model of syngamy with one-step meiosis can been extended such that the two parental haploid genomes each become a gamete alongside their one-mutant genomes, which are also recombined (randomly chosen single point crossover) with each other. Two of the four resulting haploid genomes/gametes are then chosen at random to create the cell/organism for fitness evaluation (Figure 8).

Figure 8 (right) shows the typical behaviour for various $K$. In comparison with both endomitosis and syngamy, it is found that the increased learning is beneficial for all $K>2$ (T-test, $p<0.05$), as anticipated by the results in section 3. The same general results as before are found for $N=100$ – the extra learning provides no benefit (not shown). However, greatly increasing the number of possible recombination points by using uniform crossover

[30], where each gene is swapped with equal probability, means improved performance is again seen (Figure 9) for $K>4$ (T-test, $p<0.05$, $K=6$; $p\leq0.10$, $K=10, 15$). That is, the increased potential variation in the size of the generalization (end positions) is required for the larger fitness landscape. Similarly, a significant drop in performance is seen for all cases when recombination is removed (not shown).

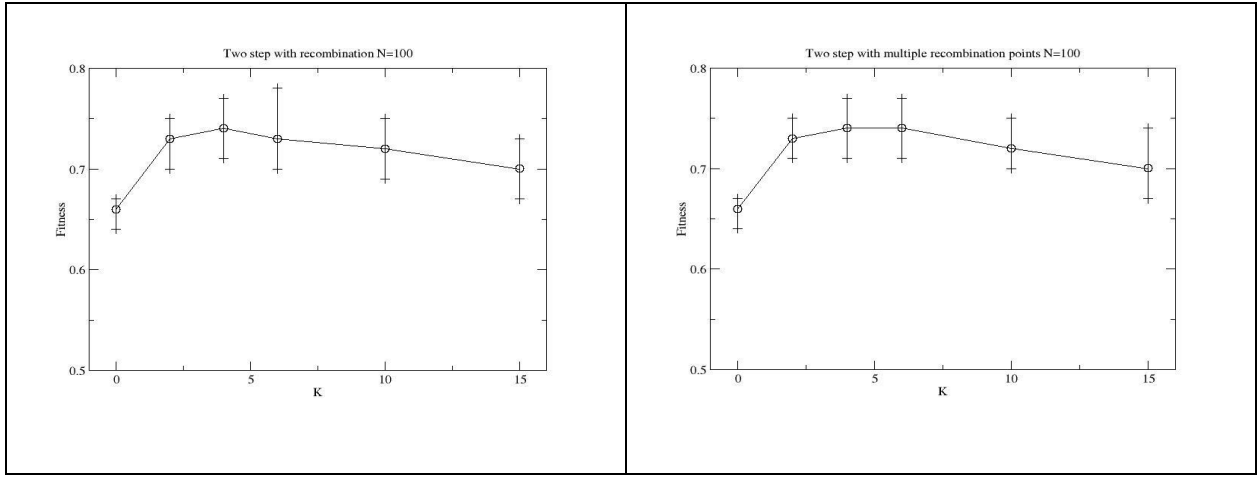

Figure 9: Showing the comparative performance of two-step meiosis with a single-point (left) and multi-point (right) recombination processes, after 50,000 generations, on landscapes of varying ruggedness ($K$) with $N=100$.

As noted above, the percentage of their lifecycle eukaryotes spend as diploids varies greatly across species. Similarly, some species alternate between being sexual and asexual, such as aphids. Following the results in Figure 3 (top), the case of half the lifecycle being spent as a haploid can be considered. Here one of the two haploids is chosen at random to make a 50% contribution to the fitness of the offspring, with the other 50% determined as the pair's average, as before. Figure 10 (left) shows how there is no significant difference in fitness for $K<10$ (T-test, $p\geq0.05$) but a significant decrease in fitness is seen for $K\geq10$ (T-test, $p<0.05$) compared to the animal-like diploid lifecycle case considered in Figure 8. Other percentages of time as a haploid have not been explored here but the results in Figure 3 suggest beneficial weightings exist. Figure 3 (bottom) also showed the potential benefits of varying the frequency of learning. Figure 10 (right) shows how varying the frequency of sexual reproduction to asexual reproduction, where the diploid is mutated once in each haploid to form an offspring in the latter, provides an increase in fitness at a ratio of asexual generations to one sexual generation of 7:1 (T-test, $p\leq0.10$) for $K=2$, with no significant change otherwise. No benefit was found for the other ratios and values of $K$ explored (not shown). However, it can also be noted that if environmental conditions vary temporally such that the underlying ruggedness of the species' fitness landscape is increased/decreased, sexual reproduction is likely to be more/less effective at that time.

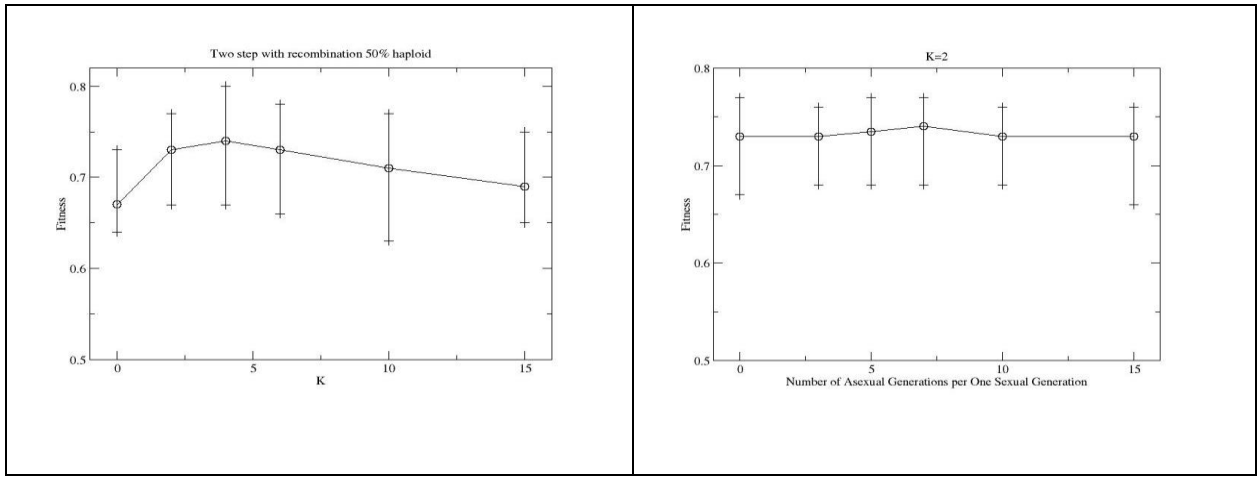

Figure 10: Showing two-step meiosis with recombination process where the lifecycle is 50% haploid and its performance, after 50,000 generations, on landscapes of varying ruggedness ($K$) with $N$=20 (left). And showing two-step meiosis with recombination with varying rounds of asexual reproduction per sexual reproduction event (right).

6. Conclusion

This paper has suggested that the haploid-diploid cycle seen in all eukaryotic sex exploits a rudimentary form of the Baldwin effect, with the diploid phase seen as the "learning" step. With this explanation for the basic cycle, the other associated phenomena such as recombination, varying the duration of the periods of haploid and diploid state, etc. can be explained as evolution tuning the amount and frequency of learning experienced by an organism. Eukaryotic evolution is seen as refining generalizations over regions of the fitness landscape to identify fit genomes, in contrast to varying the position of single points as in prokaryotic evolution. This explanation does not seemingly contradict any of the mentioned previous explanations for the various stages of eukaryotic sex, rather it presents a unifying process which underpins it and over which many other phenomena may also be occurring.

This hypothesis was based on previous work investigating the Baldwin effect which showed how the optimal amount and/or frequency of learning varied with the ruggedness of the underlying fitness landscape [6]. To demonstrate the basic idea, following [6], the well-known NK model has been extended in various ways here. It is perhaps interesting to note that, in its assuming an animal-like diploid-dominated haploid-diploid lifecycle, conditions exist in the model at $K=4$ under which both endomitosis and syngamy with a single-step meiosis are equivalent and that syngamy with a two-step meiosis and recombination is most beneficial.

A haploid-diploid cycle has not been shown beneficial in the simplest case of $K=0$. Some experimental results suggest the average degree of connectivity/epistasis in eukaryotic organisms is typically higher than in prokaryotes (eg, [20]). This offers one reason why the cycle did not evolve in prokaryotes. Further, it has been suggested that the accumulation of mitochondria – and then chloroplasts – through symbiogenesis caused an increase in the ruggedness of the fitness landscape of the resultant early eukaryote as inter-dependence became intra-dependence [5]. This can also been seen as creating/aiding the conditions under which a rudimentary Baldwin effect process would prove beneficial. Note that ploidy variation is particularly prevalent in plants (eg, [28]), where chloroplasts can be seen to further increase $K$.

The process of allele dominance can be seen as related since it can tune the amount of learning experienced on a per-gene basis. It is a further mechanism through which evolution can control the bias in the fitness value assigned to the generalization over the region of the fitness landscape defined by the two constituent haploid genome end points. Results (not shown) where which of the two genomes used to provide the fitness is chosen at random, as opposed to using their average above, gives improved performance for $K>2$ (T-test, $p<0.05$). This can be seen as the extreme case where one genome (randomly) dominates the other. Similarly, varying ploidy levels in cell types in multicellular organisms can be seen as a further mechanism by which the amount and frequency of learning is fine-tuned. That is, the ruggedness of the fitness landscape contributions for different cell types need not be uniform [11]. The simplicity of the model requires mutational differences between genomes whereas some of the other effects noted above, such as gene product concentrations, could be tuned through varying ploidy levels which may explain why higher ploidy was not beneficial here.

One of the important steps in the understanding of the evolution of sex presented here is to consider the haploid genome fitness landscape of primary importance, as opposed to that of the diploid. As noted above, it can be expected that the relationship between the fitness contributions of the haploids and that of the resulting diploid, no matter how long or often it exists, will be non-linear. Mayley [23] explored varying the degree of correlation between the underlying genome space and that of the learned phenotype space. Similar considerations should be undertaken here to extend the simple averaging approach adopted, including dominance, and to consider the dynamic nature of the landscapes caused by the pairing of genomes.

The evolution of mating types can also be seen as a way to further increase the amount of learning under the haploid-diploid cycle since the increased probability of heterozygotes under syngamy is likely to be beneficial over the scenario considered here. Todd and Miller (eg, [32]) identified the potential for the Baldwin effect to occur under the process of sexual selection, wherein variation in preferences for characteristics of (roughly equally fit) mates enables a population to escape local optima. They further suggest such mate choice may drive (sympatric) speciation. Based on the findings here, sexual selection may be seen as a mechanism through which haploid genomes are attempting to actively choose other seemingly appropriate haploids to partner with, thereby influencing/driving the changes in the shape of the underlying haploid landscape discussed above (Figure 7).

The model used here is very simple and assumes a converged population. That an equivalent asexual diploid, even one in which a small "population" of the three possible haploid genome combinations were all evaluated and the fittest chosen as the parent for the next generation if fitter, was less effective for increasing ruggedness suggests the results are not dependent upon population size (see [8] for related work using populations of many individuals). Significantly, it was shown how increasing the size of the genome space ($N$=100) decreased the benefits of the higher learning rates seen with smaller $N$ ($N$=20). Thus whilst endomitosis (to diploidy), synagmy, and syngamy with recombination were always found beneficial over the haploid case for $K$>0, the added benefit of a two-step meiosis with a single recombination point over endomitosis or syngamy was lost with $N$=100, and a greater number of recombination points were needed to retain the benefit of the latter. However, it is known that "upper and lower tolerance limits for chromosome size seem to exist for some groups of organisms" [27]. The finding here suggests why that is and, moreover, provides a subsequent reason for the maintenance of multiple chromosomes within eukaryotes. Evolution can be seen to tune chromosome length and number to make most effective use of the rudimentary learning process for the overall genome. Since significant increases in a given chromosome's size would disrupt that process, increases in overall genome size are more effectively realized through increasing the number of chromosomes since that can be seen to also increase the number of overall recombination points; dividing genes into chromosomes introduces fixed recombination points in the overall genome, in addition to varying the number of potential recombination events within each of the different chromosomes.

Future work will consider previous models reported in the literature for the evolution of eukaryotic sex in more detail with the new Baldwin effect view. In particular, the ability of a sexual species to invade an asexual species should be considered. New mechanisms within evolutionary computation also suggest themselves based on the findings here [8].